\documentclass[11pt]{article}

\usepackage[final]{acl}
\usepackage{booktabs}
\usepackage{threeparttable}
\usepackage{fontspec}
\usepackage{multirow}
\usepackage{times}
\usepackage{latexsym}
\usepackage{graphicx}
\usepackage[notransparent]{svg}
\usepackage{tikz}
\usetikzlibrary{trees}

\usepackage[T1]{fontenc}

\usepackage{float}

\usepackage{graphicx}

\usepackage[english]{babel}

\babelprovide[import]{hindi}

\usepackage[english]{babel}

\newfontfamily\hindifont{NotoSansDevanagari-Regular.ttf}[
    Script=Devanagari, 
    Scale=MatchLowercase, 
    Path=./,              
]

\newcommand{\hi}[1]{{\hindifont #1}}

\newfontfamily\ipafont[Path=./,Extension=.ttf]{CharisSIL-Regular}
\newcommand{\ipa}[1]{{\ipafont #1}}

\usepackage{xcolor}
\usepackage{enumitem}
\usepackage{tabularx}

\usepackage[verbose=silent]{microtype}

\usepackage{inconsolata}

\usepackage{hyperref}

\title{Dialect Matters: Cross-Lingual ASR Transfer \\ for Low-Resource Indic Language Varieties}

\author{Akriti Dhasmana \and Aarohi Srivastava \and David Chiang \\
    Computer Science and Engineering \\
 University of Notre Dame \\
 Notre Dame, IN, USA \\ \texttt{\{adhasman, asrivas2, dchiang\} @nd.edu}}

\begin{document}
\maketitle
\begin{abstract}

We conduct an empirical study of cross-lingual transfer using spontaneous, noisy, and code-mixed speech across a wide range of Indic dialects and language varieties. Our results indicate that although ASR performance is generally improved with reduced phylogenetic distance between languages, this factor alone does not fully explain performance in dialectal settings. Often, fine-tuning on smaller amounts of dialectal data yields performance comparable to fine-tuning on larger amounts of phylogenetically-related, high-resource standardized languages.
We also present a case study on Garhwali, a low-resource Pahari language variety, and evaluate multiple contemporary ASR models. Finally, we analyze transcription errors to examine bias toward pre-training languages, providing additional insight into challenges faced by ASR systems on dialectal and non-standardized speech.
 
\end{abstract}

\section{Introduction}

Automatic speech recognition (ASR) systems for Indic languages have made significant progress through large-scale multilingual pre-training on high-resource, standardized languages. However, these advances have largely overlooked the linguistic reality of the region, where speech is characterized by extensive dialectal variation, spontaneous and noisy recording conditions, and frequent code-switching. As a result, state-of-the-art Indic ASR models often perform poorly when applied to low-resource dialects and language varieties, even those that are closely related to  languages seen during pre-training.

Such efforts for Indic ASR typically rely on pre-training on a small set of mainstream languages such as Hindi, Marathi, and Bengali, under the assumption that phylogenetic similarity will facilitate transfer to related varieties. While this assumption provides a useful baseline in many cross-lingual settings, its sufficiency for dialectal speech remains underexplored. In practice, dialects and non-standard varieties often exhibit distinct phonological, lexical, and orthographic properties that are not well-captured by standard language data, raising questions about the effectiveness of relying solely on high-resource languages for transfer in dialectal ASR.

In this work, we examine cross-lingual transfer for Devanagari-script Indic varieties using spontaneous, noisy, and code-mixed speech. Rather than focusing solely on phylogenetic similarity, we investigate how different choices of fine-tuning data affect ASR performance on dialectal speech. Our results suggest that incorporating even limited amounts of dialectal speech during fine-tuning can be as effective as, or more effective than, relying on larger amounts of standardized language data, highlighting the role of dialectal variation in shaping transfer behavior.

To complement our cross-lingual analysis, we present a case study on Garhwali, a low-resource Pahari language variety that has been largely absent from prior ASR research. We evaluate several self-supervised speech models on Garhwali and perform a detailed error analysis to characterize the challenges posed by dialectal variation, code-mixing, and model bias toward pre-training languages. This case study provides concrete insights into the limitations of current Indic ASR models when applied to low-resource language varieties.

\paragraph{Contributions} In this paper, we make three contributions:
\begin{enumerate}
    \item We provide a comprehensive empirical analysis of cross-lingual transfer for low-resource Devanagari-script Indic dialects and show that fine-tuning on other dialects is often more effective than relying on closely-related high-resource standardized languages.
    \item We present the first detailed ASR study for Garhwali, including model evaluation and qualitative and quantitative error analysis.
    \item We introduce a diagnostic approach for quantifying bias toward pre-training languages in dialectal ASR, enabling systematic analysis of model behavior on non-standardized and code-mixed speech.
\end{enumerate}

\section{Related Work}
Recent work has highlighted systematic performance gaps between standard language varieties and regional or minority dialects across NLP and speech technologies. \citet{kantharuban2023quantifying} provide a large-scale evaluation of state-of-the-art models for machine translation and automatic speech recognition across regional dialects of multiple high- and low-resource languages, showing that dialectal performance disparities are widespread and variably correlated with linguistic, social, and data-related factors. Complementary to this line of work, \citet{blaschke2025standardtodialecttransfertrendsdiffer} study standard-to-dialect transfer in spoken and written settings for German, demonstrating that speech-based models are more robust to dialectal variation than text-based or cascaded approaches, particularly when orthographic normalization is involved.

In the Indic context, prior work has investigated dialectal variation primarily through multilingual or language-specific ASR systems. For example, \cite{kumar25_interspeech} evaluate a multilingual dialect identification and ASR pipeline across 33 dialects of eight Indic languages using read speech, while other efforts have focused on building comprehensive ASR toolkits for standardized Indic languages \cite{chadha2022vakyanshasrtoolkitlow}. However, cross-dialectal and cross-lingual transfer for Indic ASR—especially in zero-shot settings and on spontaneous, non-standardized speech—has not yet been systematically examined. Our work addresses this gap by analyzing cross-lingual transfer behavior and dialectal bias in low-resource Indic language varieties.

The effects of orthographic irregularity and transcription variability on ASR performance have been examined in other language contexts, including cross-family settings \cite{taguchi-chiang-2024-language} and Swiss German dialectal ASR \cite{nigmatulina-etal-2020-asr}. We build on these findings to analyze how similar forms of orthographic variation impact ASR performance for Indic dialects.

Prior work on speech technology for Garhwali has primarily focused on language identification \cite{Gusain2023AutomaticLI} and the creation of domain-specific datasets, such as for agriculture-related applications \cite{article}. To the best of our knowledge, no prior work has involved training or evaluating an ASR model for Garhwali.

The widespread presence of multiple language varieties in India, shaped by historical and sociolinguistic factors, has led to extensive code-mixing in everyday speech, posing additional challenges for ASR systems. While code-mixed speech has been studied in specific settings such as Hindi-Marathi ASR \cite{10835062}, we provide a systematic way to quantify transcription errors arising from code-mixing in dialectal ASR.

\section{Languages and Data}
\begin{figure*}[t]
  \centering
  \resizebox{\textwidth}{!}{
\begin{tikzpicture}[
  level distance=25mm,
  sibling distance=50mm,        
  edge from parent/.style={draw, thick},
  every node/.style={font=\large, align=center},
  seen/.style={draw=blue!70, thick, rounded corners, fill=blue!10, inner sep=4pt, anchor=north},
  unseen/.style={draw=black!40, rounded corners, fill=gray!10, inner sep=4pt, anchor=north},
  root/.style={font=\bfseries\large}
]

\tikzstyle{no edge from parent}=[
    edge from parent path={}
]

    \node[level distance=10mm] {Indo-Aryan Languages}
        child[level distance=10mm] { node {Continental Indo-Aryan Languages}
                child[level distance=10mm, sibling distance=170mm] { node { Indo-Aryan Eastern Zone} child[level distance=10mm] { node[unseen] {Halbi\\WER 0.87}}}
                child[level distance=10mm, sibling distance=120mm] { node {Midlands Indo-Aryan}
                child[level distance=10mm, sibling distance=65mm] { node {Indo-Aryan Northern zone}
            child[level distance=30mm, sibling distance=35mm] { node {Central Pahari}
              child[grow=down, level distance=7mm, sibling distance=0mm] { node[unseen] {Kumaoni\\WER 0.60}
                child[no edge from parent] { node[unseen] {Garhwali\\WER 0.82} }
              }
            }
            child[level distance=30mm, sibling distance=20mm] { node {Eastern Pahari}
              child[grow=down, level distance=7mm, sibling distance=0mm] { node[seen] {Nepali\\WER 0.96} }
            }
          }
          child [level distance=10mm,sibling distance=50mm]{ node {Shaurasenic}
            child[level distance=10mm, sibling distance=35mm] { node {Bihari Group} 
              child[grow=down, level distance=7mm, sibling distance=5mm] { node[seen] {Bhojpuri\\WER 0.73} 
                child[no edge from parent] { node[seen] {Maithili\\WER 0.70} 
                  child[no edge from parent] { node[unseen] {Bajjika\\WER 0.78} 
                    child[no edge from parent] { node[unseen] {Angika\\WER 0.73} 
                      child[no edge from parent] { node[unseen] {Magadhi\\WER 0.70} 
                        child[no edge from parent] { node[unseen] {Surjapuri\\WER 0.88} 
                          child[no edge from parent] { node[unseen] {Thethi\\WER 0.71}
                                child[no edge from parent]{ node[unseen]{Sadri\\WER 0.74}}
                          }
                        }
                      }
                    }
                  }
                }
              }
            }
            child[level distance=10mm, sibling distance=35mm] { node {Eastern Hindi}
              child[grow=down, level distance=7mm, sibling distance=35mm] { node[seen] {Chhattisgarhi\\WER 0.76}
                child[no edge from parent] { node[unseen] {Surgujia\\WER 0.75}
                  child[no edge from parent] { node[unseen] {Awadhi\\WER 0.77} }
                }
              }
            }
            child[level distance=10mm, sibling distance=30mm] {
            node {Central Zone}
           child[level distance=10mm, sibling distance=35mm] { node {Western Hindi}
              child[grow=down, level distance=7mm, sibling distance=35mm] { node[seen] {Hindi\\WER 0.50}
                child[no edge from parent] { node[unseen] {Khariboli\\WER 0.61}
                  child[no edge from parent] { node[unseen] {Haryanvi\\WER 0.77}
                    child[no edge from parent] { node[unseen] {Bundeli\\WER 0.65} }
                  }
                }
              }
            }
            }
          }
          child[sibling distance=79mm] { node {Apabhramsic}
            child[level distance=10mm, sibling distance=15mm] { node {Gujarati-Rajasthani Group} 
              child[level distance=10mm, sibling distance=15mm] { node {Rajasthani Group}
              child[grow=down, level distance=7mm, sibling distance=0mm] { node[seen] {Rajasthani\\WER 0.76}
                child[no edge from parent] { node[unseen] {Marwari\\WER 0.74}
                  child[no edge from parent] { node[unseen] {Jaipuri\\WER 0.53}}
                  }
                }
              }
            }
          }
          child[sibling distance=60mm]{ node {Indo-Aryan Southern Zone}
            child[level distance=10mm, sibling distance=10mm] { node {Marathic}
            child[level distance=10mm, sibling distance=10mm] { node {Marathi-Konkani}  child[level distance=10mm, sibling distance=35mm] { node {Old-Modern-Marathi}
            child[grow=down, level distance=7mm, sibling distance=10mm] { node[seen] {Marathi\\WER 0.87} }
          }
          child[level distance=10mm, sibling distance=35mm] { node {Goan Konkani}
            child[grow=down, level distance=7mm, sibling distance=0mm] { node[seen] {Konkani\\WER 0.92} 
            child[no edge from parent] { node[unseen] {Malvani\\WER 0.88}}
            }
          }
          }}}
        }
};

\end{tikzpicture}
}
\caption{Subset of the Indo-European language family tree showing Devanagari-script Indic languages in the VAANI dataset, based on Glottolog \cite{indo1320}. Languages are annotated with WER for \texttt{IndicWav2Vec-Hindi}. Blue highlights indicate languages used during pre-training.}
\label{fig:language_tree}
\end{figure*}
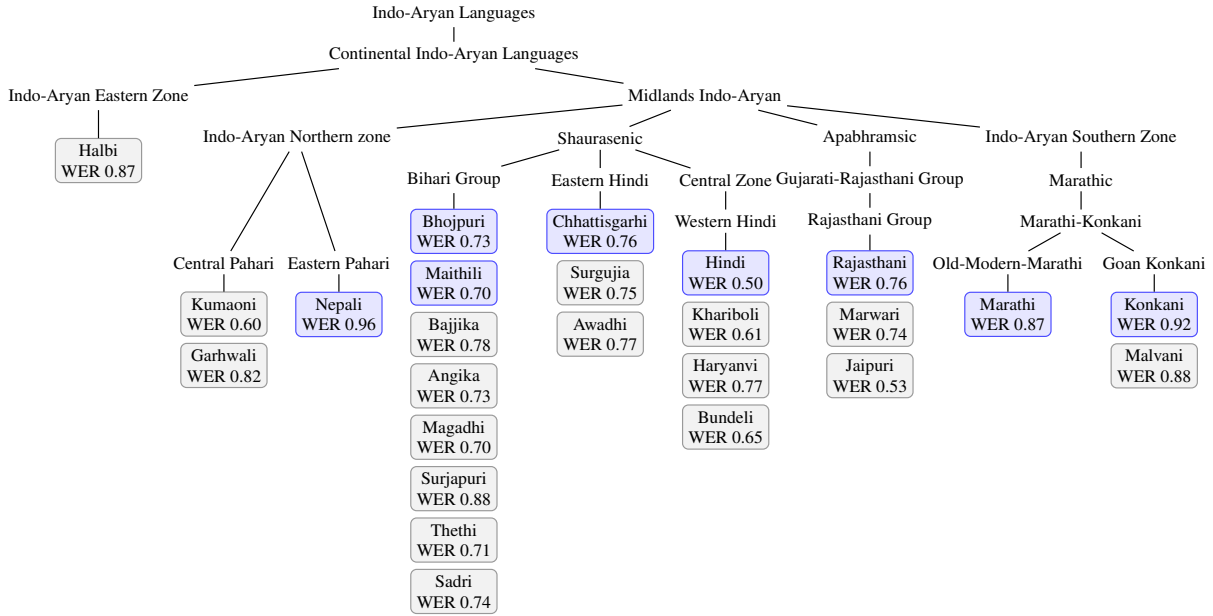

The Indic linguistic landscape is characterized by a high degree of diversity, encompassing hundreds of languages and dialects with varying levels of standardization and resource availability. Many widely spoken languages coexist with numerous regional dialects that differ substantially in phonology, morphology, and lexicon, despite phylogenetic relatedness \cite{massica_1993}. In everyday use, speakers frequently engage in code-mixing, particularly with English, and in spontaneous and acoustically noisy environments. These factors pose major challenges for ASR systems trained primarily on clean, standardized, and monolingual speech \cite{diwan21_interspeech}. We present a phylogenetic tree of the languages included in our experiments in Figure~\ref{fig:language_tree}. 

Most existing ASR resources for Indic languages focus on a small subset of high-resource, standardized languages, leaving dialectal varieties underrepresented or entirely absent. To address this gap, we fine-tune and evaluate our models on the VAANI dataset \citep{VAANI2025}.\footnote{\url{https://huggingface.co/datasets/ARTPARK-IISc/VAANI}} VAANI captures the rich linguistic diversity present across speakers of Indic languages, including different accents, grammatical variation, loan words, and code-switching patterns, all of which are vital to include in the context of Indic ASR.

VAANI consists of spontaneous speech collected by prompting participants to describe images in their local dialect. The dataset contains over 150,000 hours of speech, of which approximately 10\% is transcribed, and covers 156,534 speakers from 773 districts across India. As Indic languages are written in multiple scripts, we focus our analysis on language varieties written in the Devanagari script. Due to the broad coverage of dialects and language varieties, the amount of available data varies substantially across languages in the dataset; we account for this variability as much as possible in our experimental design.

Beyond conducting cross-lingual experiments across Indic language varieties, we also conduct a detailed analysis of ASR for one nonstandard language variety. We select \emph{Garhwali} for this case study.\footnote{According to the Post-1971 Census of India, these languages were recognized as ``mother tongues'' and were designated as dialects of Hindi \cite{KHUBCHANDANI1991265}; however, some linguists argue that Garhwali and other Pahari languages are distinct languages \cite{Gusain2023AutomaticLI}.} Garhwali is a language that belongs to the Pahari \emph{subgroup} \footnote{We refer to the lowest non-leaf nodes in a phylogenetic tree as \emph{subgroups}.} spoken in the Himalayan region, particularly in the state of Uttarakhand \cite{Gusain2023AutomaticLI}.

\section{Experiments and Results}
In our experiments, we aim to answer the following questions to analyze broader patterns in Indic ASR:
\begin{enumerate}[nosep,itemindent=1em]
    \item[RQ1.] How does a strong Indic ASR baseline perform across a diverse set of low-resource language varieties, particularly those not seen during model pre-training?
    \item[RQ2.] To what extent does orthographic variability correlate with ASR performance in Indic language varieties?
    \item[RQ3.] How does cross-lingual transfer for Indic ASR vary with phylogenetic distance, and does this relationship hold for dialectal speech?
\end{enumerate}
    In addition, we conduct a focused case study on Garhwali to characterize systematic error patterns in dialectal ASR. We analyze how systemic bias towards pre-training languages (Hindi) manifest in transcriptions generated by ASR models fine-tuned on a dialect (Garhwali).
\paragraph{Metrics}
We employ two standard metrics, word error rate (WER) and character error rate (CER), to evaluate the generated transcriptions. Both metrics are calculated based on Levenshtein distance, representing the minimum number of insertions, deletions, and substitutions required to align the hypothesis with the reference text. We report WER in our main results and include CER in the appendix.

\subsection{RQ1: Baseline Assessment for Indic ASR}
\paragraph{Setup} We begin by assessing the performance of a state-of-the-art Indic ASR model, \texttt{IndicWav2Vec} \cite{javed2021buildingasrsystemsbillion}, on language varieties present in the VAANI dataset. \texttt{IndicWav2Vec} is based on the Wav2Vec 2.0 architecture \cite{baevski2020wav2vec20frameworkselfsupervised}, and is pre-trained on 17,000 hours of unlabeled \emph{clean} speech data from YouTube, as well as Newsonair data curated from radio channels, across 40 Indic languages. In our experiments, we employ \texttt{IndicWav2Vec-Hindi}\footnote{\url{https://huggingface.co/ai4bharat/IndicWav2Vec-Hindindi}} (\texttt{IndicWav2Vec} fine-tuned on Hindi), since pre-trained ASR models largely learn language-agnostic acoustic representations during pre-training and require language-specific fine-tuning to map these representations to text transcriptions.
\begin{figure*}[ht!]
\includegraphics[width=\textwidth]{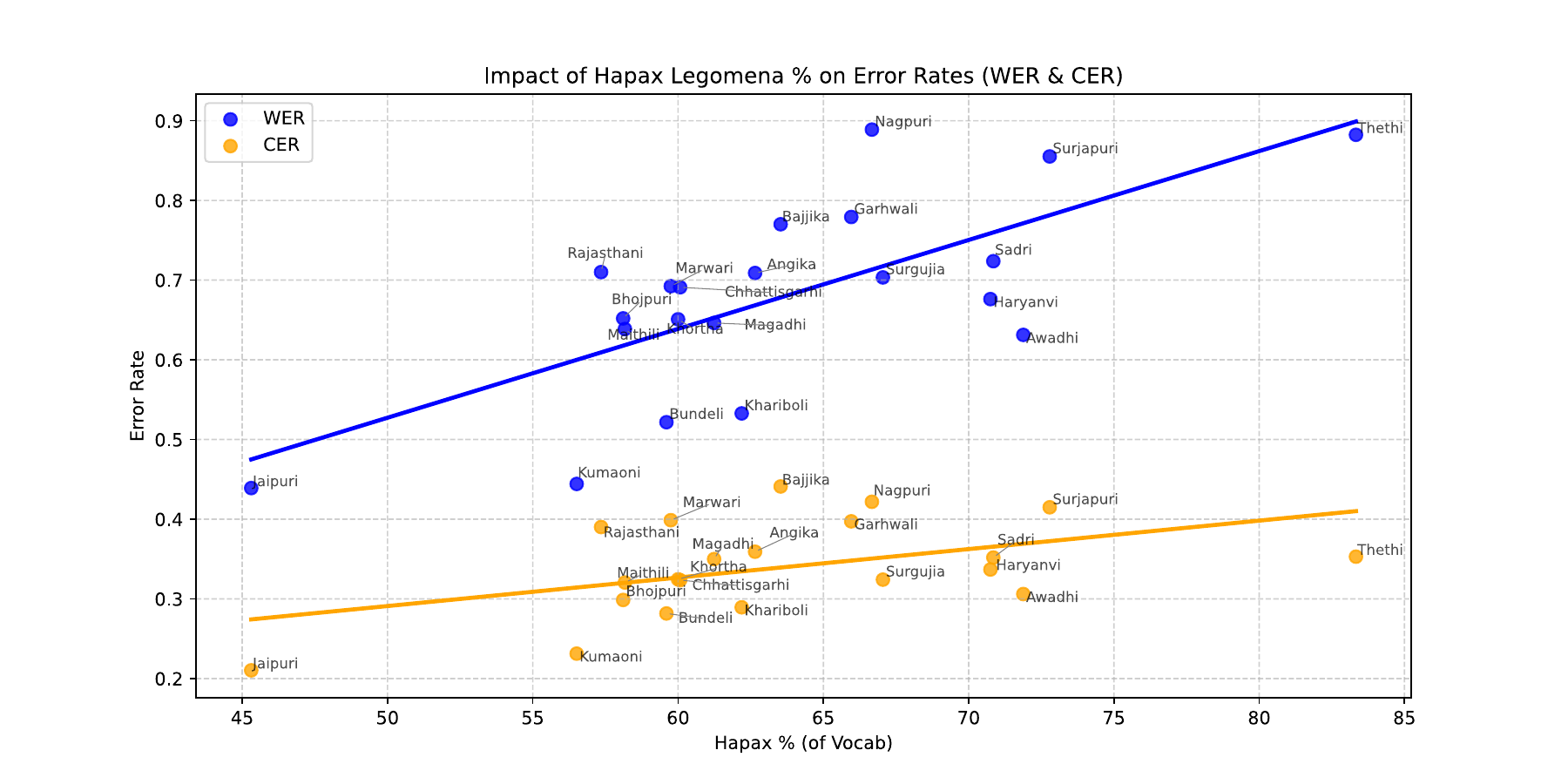}
\caption{Hapax legomena (once-seen word types) \% per language in the test split of VAANI plotted against WER (Pearson's $\rho = 0.705$, $p = 4 \times 10^{-4}$) and CER (\emph{not significant})
using \texttt{IndicWav2Vec-Hindi}.}
\label{fig:ImpactofHapax}
\end{figure*}

\paragraph{Results}
We evaluate \texttt{IndicWav2Vec-Hindi} on test samples for all 30 Devanagari-script varieties in VAANI (up to 1 hour each) and report these results as part of a phylogenetic tree in Figure \ref{fig:language_tree}. We also divide the results into two tables depending on whether the language was used to pre-train \texttt{IndicWav2Vec} (Table~\ref{tab:seen_languages_split}) or it was not explicitly seen by the model (Table~\ref{tab:unseen_languages_split}). These results show that, despite being fine-tuned on Hindi, \texttt{IndicWav2Vec-Hindi} achieves a best-case word error rate of 50.4\% on VAANI Hindi test speech (one hour). Similarly high error rates are observed for several languages seen during pre-training, indicating that pre-training alone does not ensure robust performance on spontaneous and noisy speech, even for languages included in the pre-training corpus. We also observe substantial variation in performance across dialects and closely-related language varieties. These observations suggest that other factors, such as geographical proximity or phonological feature similarity, could have an impact on performance, motivating our subsequent analyses.

\subsection{RQ2: Orthographic Consistency}

\paragraph{Setup}Orthographic consistency is an important factor in achieving high ASR performance, as inconsistent spellings introduce additional variability that can increase modeling and decoding errors \cite{nigmatulina-etal-2020-asr}. This challenge is particularly pronounced for Indic languages that encompass multiple dialects where orthographic conventions are not fixed. To quantify orthographic irregularity in the data, we measure the frequency of each word type in the transcripts and analyze the proportion of tokens that occur only once, along with the type-to-token ratio. We then examine how these measures correlate with ASR performance.

\paragraph{Results}
We observe a clear trend between orthographic variability, as measured by the proportion of \emph{hapax legomena} (unique words) in the test set, and ASR performance. Figure~\ref{fig:ImpactofHapax} shows that languages with a higher percentage of unique words exhibit higher WERs when evaluated with \texttt{IndicWav2Vec-Hindi} (Pearson's $\rho = 0.705$, $p = 4 \times 10^{-4}$), indicating increased difficulty for languages with less consistent orthographic conventions (e.g., Thethi, Surjapuri). A similar but weaker trend is observed for CER.

Additional evidence is provided by the type-token statistics reported in Table~\ref{tab:hapax_analysis}. Several languages (e.g., Gondi, Thethi) exhibit a high number of distinct word types relative to the total number of tokens, resulting in elevated type-to-token ratios. These measures reflect greater orthographic and lexical variability and are associated with higher ASR error rates. Taken together, these results suggest that orthographic inconsistency is an important contributing factor to reduced ASR performance in nonstandard Indic varieties.

\begin{figure*}[t]
\hspace*{-7mm}\includegraphics[width=1.13\textwidth]{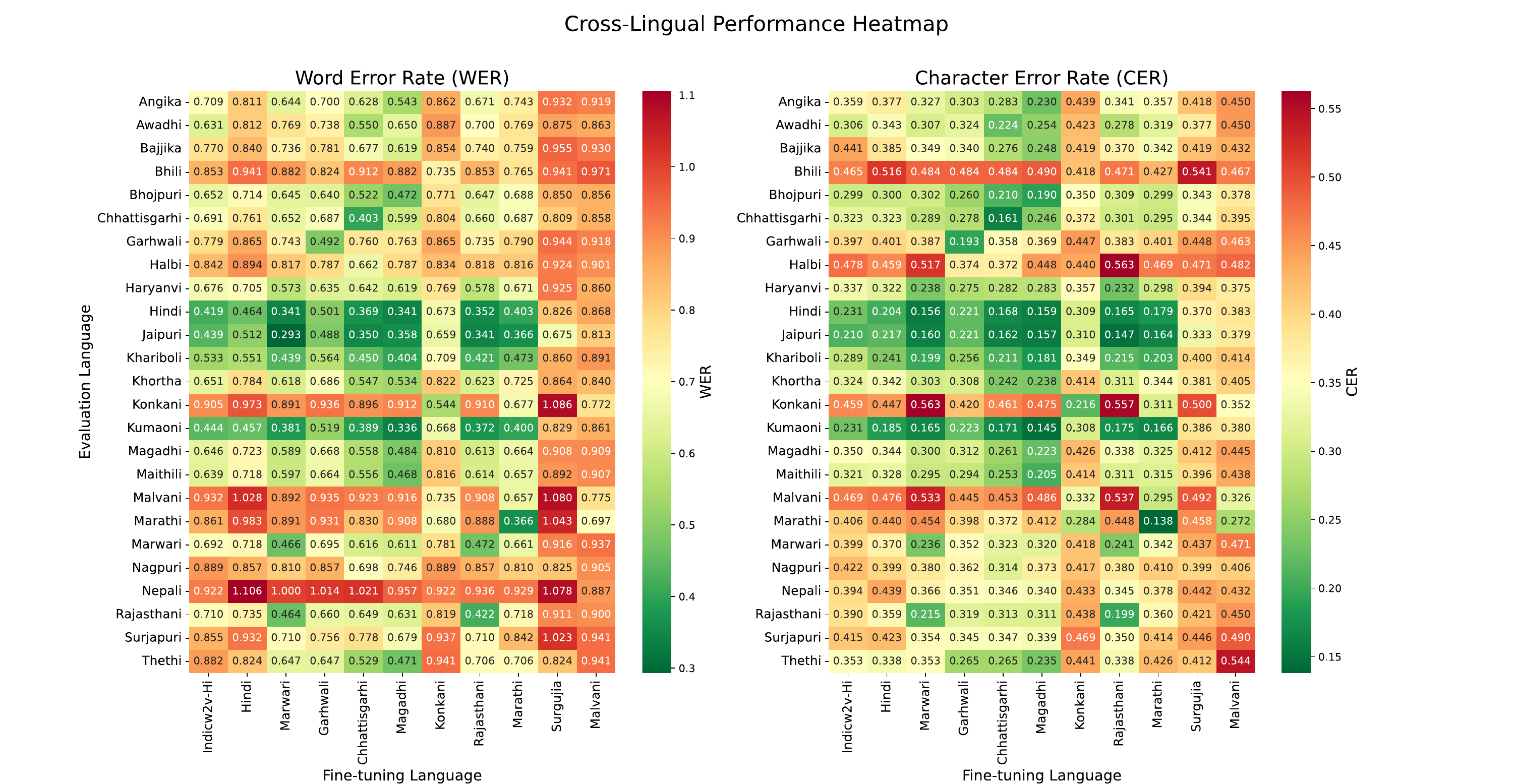}
\caption{Cross-Lingual Performance of \texttt{w2vBERT} models fine-tuned on 1 to 7 hours of data per language vs. off-the-shelf \texttt{IndicWav2Vec-Hindi}.}
\label{fig:heatmap}
\end{figure*}

\subsection{RQ3: Cross-Lingual Transfer}
We wanted to examine whether there are any performance gains from cross-dialectal fine-tuning in zero-shot settings. For each of these experiments, we fine-tune \texttt{w2v-bert-2.0}\footnote{\url{https://huggingface.co/facebook/w2v-bert-2.0}} (selected through preliminary comparison of models) on 1 to 7 hours of speech, depending on the availability. Each model was then evaluated on the test set of all Devanagari script languages available from VAANI. We compute correlations between the error rate and phylogenetic distance under two conditions: across all evaluation languages and restricted to nonstandard test varieties. We additionally measure the association between the error rate and whether the fine-tuning language represents dialectal or standard speech. We consider the following languages from the dataset to be dialects or nonstandard varieties: Angika, Awadhi, Bajjika, Bhili, Chhattisgarhi, Garhwali, Halbi, Haryanvi, Jaipuri, Khariboli, Khortha, Kumaoni, Magadhi, Malvani, Marwari, Nagpuri, Surjapuri, and Thethi.\footnote{None of these languages belong to the Eighth Schedule to the Indian Constitution listing the officially recognized languages (\url{https://en.wikipedia.org/wiki/Eighth_Schedule_to_the_Constitution_of_India}).}

\paragraph{Results}
Figure~\ref{fig:heatmap} summarizes cross-lingual transfer performance for \texttt{w2vBERT} models fine-tuned on 1 to 7 hours of data per language and evaluated in a zero-shot setting across a wide range of Indic language varieties, providing a holistic view of how ASR performance varies as a function of both the fine-tuning and the evaluation languages.

Across all evaluation languages, we observe a clear association between phylogenetic distance and ASR performance: larger distances between the fine-tuning and evaluation languages generally correspond to higher word error rates (Spearman's $\rho = 0.333$, $p = 1.11 \times 10^{-7}$). This result aligns with prior findings in cross-lingual ASR and confirms that phylogenetic relatedness provides a useful baseline for transfer across Indic languages. However, this aggregate trend does not fully characterize model behavior when evaluation is restricted to dialectal speech.

\subsubsection{Dialectal Transfer Effects}
When focusing specifically on dialects and nonstandard test varieties, the heatmap reveals additional structure that is not explained by phylogenetic distance alone. Although the association between phylogenetic distance and word error rate remains statistically significant in this setting (Spearman's $\rho = 0.274$, $p = 3.604 \times 10^{-4}$), several consistent deviations from this trend are observed. In particular, models fine-tuned on nonstandard varieties frequently outperform models fine-tuned on phylogenetically closer but standard, higher-resource languages.

One representative example from Figure~\ref{fig:heatmap} is the strong transfer from Marwari (a dialect of Rajasthani) to Kumaoni (a Pahari language variety), even though the two belong to distinct phylogenetic subgroups and are geographical distant. Another example is the consistently competitive performance of the models fine-tuned on Marwari and Magadhi across multiple evaluation language varieties, often outperforming models fine-tuned on higher-resource standardized languages such as Hindi, Marathi, and Rajasthani. Notably, this trend holds despite the fact that the Magadhi model is trained on less data (5 vs. 7 hours). These cases illustrate that while fine-tuning on a closely related high-resource standardized language may be the most natural strategy, it is not necessarily the most effective choice for transfer to dialects and related language varieties.

To further examine this pattern, we analyze whether the dialectal status of the fine-tuning language itself is associated with performance on unseen dialects. We find a statistically significant trend indicating that fine-tuning on dialectal speech is associated with lower WER on dialectal evaluation sets (point-biserial correlation $r_{pb} = -0.196$, $p = 5.79 \times 10^{-3}$). This trend holds even when the available dialectal training data is smaller than that of the corresponding standardized languages, suggesting that dialectal fine-tuning captures information that is not adequately represented by mainstream language data alone.

\paragraph{Controlled In-Group Transfer Experiments}

\begin{table}[!htp]
    \centering
    \small
    \begin{tabularx}{\columnwidth}{@{}Xccc@{}}
    \toprule
    \textbf{Fine-Tuning Language} &
    \textbf{Hours} &
    \multicolumn{2}{c}{\textbf{Jaipuri}} \\
    &&\textbf{WER} & \textbf{CER} \\
    \midrule
    \textbf{Rajasthani} + Marwari & 16.8 & 0.309 & \textbf{0.151} \\
    Marwari (\emph{dialect only}) & \phantom{1}7.2& \textbf{0.301} & 0.162 \\
    \bottomrule
    \end{tabularx}
    \caption{Effect of fine-tuning language choice on Jaipuri ASR (Rajasthani subgroup). Mainstream language is in bold.}
    \label{tab:rajasthani}
\end{table}

\begin{table}[t]
    \centering
    \small
    \begin{tabularx}{\columnwidth}{@{}Xccc@{}}
    \toprule
    \textbf{Fine-Tuning Language} & \textbf{Hours} & \multicolumn{2}{c}{\textbf{Thethi}} \\
    && \textbf{WER} & \textbf{CER} \\
    \midrule
    Angika, Bajjika, \textbf{Bhojpuri}, Khortha, Magadhi, \textbf{Maithili}, Sadri & 50.6 & 0.412 & \textbf{0.132} \\
    Angika, Bajjika, Khortha, Magadhi, Sadri (\emph{dialects only}) & 15.6 &\textbf{0.353} & 0.147 \\
    \bottomrule
    \end{tabularx}
    \caption{Effect of mainstream language inclusion on Thethi ASR (Bihari subgroup). Mainstream language is in bold.}
    \label{tab:bihari_model_performance}
\end{table}

To isolate the effect of dialectal vs. standard fine-tuning data from broader cross-lingual trends, we conduct controlled transfer experiments within individual phylogenetic subgroups. These experiments compare models fine-tuned on dialectal varieties against models fine-tuned on higher-resource standardized languages within the same subgroup.
\begin{table}
    \centering
    \small
    \begin{tabularx}{\columnwidth}{@{}Xccc@{}}
    \toprule
    \textbf{Fine-Tuning Language} & 
    \textbf{Hours} &
    \multicolumn{2}{c}{\textbf{Haryvani}} \\
    &&\textbf{WER} & \textbf{CER} \\
    \midrule
    \textbf{Hindi} + Bundeli + Khariboli & 101.7 & 0.930 & 1.452 \\
    Bundeli + Khariboli (\emph{dialects only}) & \phantom{10}1.7 & \textbf{0.513} & \textbf{0.267} \\
    \bottomrule
    \end{tabularx}
    \caption{Impact of fine-tuning (FT) languages on Haryanvi ASR (Western Hindi subgroup). Mainstream language is in bold.}
    \label{tab:westernhindi}
\end{table}
Across the Rajasthani, Western Hindi, and Bihari subgroups, we consistently observe that fine-tuning on small amounts of dialectal speech yields performance comparable to or better than fine-tuning on substantially larger amounts of standardized language data. For example, within the Rajasthani subgroup (Table~\ref{tab:rajasthani}), a model fine-tuned solely on a dialectal variety (Marwari) achieves comparable performance on an unseen dialect (Jaipuri) to a model trained on a combination of dialectal and macro-language data. Similar trends are observed in the Western Hindi and Bihari subgroups (Tables~\ref{tab:westernhindi} and~\ref{tab:bihari_model_performance}), where excluding higher-resource mainstream languages does not degrade, and in some cases improves, performance on dialectal evaluation sets.

Taken together, the heatmap analysis and controlled in-group experiments demonstrate that while phylogenetic similarity provides a useful starting point for cross-lingual transfer, effective ASR for dialectal speech depends critically on the inclusion of dialectal training data itself.

\section{Garhwali ASR Case Study}

\subsection{Baseline Selection}
\paragraph{Setup}In order to determine the ideal model architecture for Garhwali, we compared several widely-used ASR models: 
\begin{itemize}[nosep]
    \item Wav2Vec2 \cite{baevski2020wav2vec20frameworkselfsupervised}
    \item HuBERT \cite{hsu2021hubertselfsupervisedspeechrepresentation}
    \item XLS-R \cite{babu2021xlsrselfsupervisedcrosslingualspeech}
    \item Whisper \cite{radford2022robustspeechrecognitionlargescale}
    \item w2vBERT \cite{chung2021w2vbertcombiningcontrastivelearning}
\end{itemize}
We fine-tune each of these models on the Garhwali subset of VAANI, using an 87\%--6\%--7\% train--test--validation ratio. 

We fine-tuned and evaluated several self-supervised speech models on the Garhwali subset of the VAANI dataset. Table \ref{tab:garhwali_models_performance} reports performance across these models.

\paragraph{Results}
Among the evaluated configurations, the \texttt{w2vBERT}-based model achieves the lowest error rates and is therefore selected for subsequent analysis. We compare this model against \texttt{IndicWav2Vec-Hindi} (used in our cross-lingual experiments). Our fine-tuned model performs better than the \texttt{IndicWav2Vec-Hindi} model; however, the resulting error rates remain high, indicating that Garhwali ASR remains challenging even with dialect-specific fine-tuning. We therefore focus our error analysis on this best-performing configuration to better understand the remaining sources of error.

\begin{table}
\centering
\small
\setlength{\tabcolsep}{4pt}
\begin{tabular}{@{}lccc@{}}
\toprule
\textbf{Model} & \textbf{Metric Optimized} & \textbf{WER} & \textbf{CER} \\
\midrule
\texttt{XLS-R}         & WER & 0.650 & 0.270 \\
\texttt{wav2vec2-BERT} & CER & \textbf{0.493} & \textbf{0.193} \\
\texttt{Whisper-small} & CER & 0.629 & 0.650 \\
\texttt{HuBERT}        & CER & 0.515 & 0.199 \\
\bottomrule
\end{tabular}
\caption{Comparison of speech models fine-tuned on Garhwali. Training details are in Table \ref{tab:app_garhwali_models_performance}.}
\label{tab:garhwali_models_performance}
\end{table}

\subsection{Error Analysis}
We conduct an extensive error analysis of the best fine-tuned model on Garhwali to identify the following:

\paragraph{Inconsistent English Transliteration}
 English text is marked in VAANI's annotations; we assess ASR output specifically on these segments. A prominent source of error arises from inconsistent transliteration of English words into the Devanagari script. Due to frequent code-switching in spontaneous Indic speech, English words commonly appear in the ground-truth transcripts. However, the absence of standardized conventions for English-to-Devanagari transliteration, combined with imperfect phoneme-to-grapheme mappings, results in substantial variation in the labels. For example, the training data contains multiple spellings for the English word \textit{photo}, such as \hi{फोटु} and \hi{फाटो}, reflecting pronunciation differences.

This variability introduces ambiguity during training and evaluation, complicating the model's ability to learn consistent lexical representations. Whether standardizing transliterated forms would improve dialectal ASR performance or instead remove linguistically meaningful variation remains an open question.
\begin{table}[t]
    \centering
    \small
    \setlength{\tabcolsep}{1pt}
    \newcolumntype{Y}{>{\centering\arraybackslash}X}
    \begin{tabularx}
    {\columnwidth}{@{} l Y Y Y Y @{}}
        \toprule
        \textbf{Model} & \textbf{\# Non-Hi} & \textbf{Correct} & \textbf{To Hi} & \textbf{To Wrong} \\
        \midrule
        \texttt{wav2vec2-BERT}
           & 1873 & 34.8\% & 23.3\% & 38.8\% \\
        \texttt{Indicw2v2}
           & 2109 & 4.2\% & 37.3\% & 42.9\% \\
        \bottomrule
    \end{tabularx}
    \caption{Comparison of \texttt{w2vBERT} model fine-tuned on Garhwali against \texttt{IndicWav2Vec-Hindi} on Garhwali non-Hindi word handling. Key: Hi = Hindi.}
    \label{tab:non_hindi_analysis}
\end{table}

\paragraph{Bias Towards Hindi from Pre-training}
We identify non-Hindi words in the ground-truth transcripts using \texttt{Hindi-HunSpell},\footnote{\url{https://github.com/Shreeshrii/hindi-hunspell}} a Hindi spell-checker. These includes Garhwali words, as well as transliterated English words; we separate the English words (see below). We then assess which non-Hindi words are preserved in the generated transcription vs. which are transcribed to a Hindi word.
We analyze commonly omitted, added, and substituted characters in the generated transcriptions, qualitatively identifying systematic errors and highlighting areas for future improvement.
We also observe systematic bias towards Hindi arising from model pre-training. Many errors involve Garhwali words or transliterated English terms being incorrectly normalized to valid Hindi words.

To quantify this effect, we identify non-Hindi words in the ground-truth transcripts, including Garhwali-specific vocabulary and transliterated English terms, using a Hindi spell checker. We then track three outcomes: non-Hindi words correctly preserved (\texttt{Correct}), non-Hindi words converted into valid Hindi words (\texttt{No-Hi$\rightarrow$Hi}), and non-Hindi words converted into incorrect forms (\texttt{No-Hi$\rightarrow$Wrong}). Note that the total number of non-Hindi words in the ground truth transcripts differs slightly between the two models due to preprocessing and tokenization; however, the observed trends are robust to these discrepancies.

As shown in Table~\ref{tab:non_hindi_analysis}, the \texttt{w2vBERT} model fine-tuned on Garhwali converts approximately 22.9\% of non-Hindi terms into Hindi words, while preserving roughly one-third of such terms. In contrast, \texttt{IndicWav2Vec-Hindi} preserves fewer than 5\% of non-Hindi words, converting or distorting the majority. These results indicate that ASR pre-training on mainstream Indic languages can substantially hamper a model's ability to retain dialect-specific information.

\begin{table}
\newcommand{\ins}{$\epsilon$}
\newcommand{\del}{$\epsilon$}
\newcommand{\spc}{\textvisiblespace}
    \centering
    \small
    \renewcommand{\tabcolsep}{0.6em}
    \begin{tabular}{@{}l@{ }l l@{ }l r | l@{ }l l@{ }l r@{}}
        \toprule
        \multicolumn{2}{c}{\textbf{Actual}} & 
        \multicolumn{2}{c}{\textbf{Predicted}} & 
        \textbf{\#} &
        \multicolumn{2}{c}{\textbf{Actual}} & 
        \multicolumn{2}{c}{\textbf{Predicted}} & 
        \textbf{\#} \\
        \textbf{\scriptsize{Deva}} & \textbf{\scriptsize{IPA}} & \textbf{\scriptsize{Deva}} & \textbf{\scriptsize{IPA}} & & 
        \textbf{\scriptsize{Deva}} & \textbf{\scriptsize{IPA}} & \textbf{\scriptsize{Deva}} & \textbf{\scriptsize{IPA}} & \\
        \midrule
        \spc & & \del & & 438 & \ins & & \hi{ं} & \ipa{/n/} & 68 \\
        \ins & & \spc & & 289 & \hi{ु} & \ipa{/ʊ/} & \hi{ू} & \ipa{/uː/} & 66 \\
        \hi{ा} & \ipa{/aː/} & \del & & 175 & \hi{ी} & \ipa{/iː/} & \hi{ि} & \ipa{/ɪ/} & 55 \\
        \hi{्} & & \del & & 170 & \hi{े} & \ipa{/eː/} & \del & & 51 \\
        \ins & & \hi{्} & & 163 & \hi{ु} & \ipa{/ʊ/} & \del & & 51 \\
        \hi{ं} & \ipa{/n/} & \del & & 145 & \hi{ि} & \ipa{/ɪ/} & \del & & 48 \\
        \ins & & \hi{ा} & \ipa{/aː/} & 126 & \hi{ह} & \ipa{/ɦ/} & \del & & 46 \\
        \hi{ू} & \ipa{/uː/} & \hi{ु} & \ipa{/ʊ/} & 94 & \ins & & \hi{े} & \ipa{/eː/} & 45 \\
        \hi{ि} & \ipa{/ɪ/} & \hi{ी} & \ipa{/iː/} & 90 & \hi{ै} & \ipa{/ɛː/} & \hi{े} & \ipa{/eː/} & 44 \\
        \ins & & \hi{य} & \ipa{/j/} & 81 & \ins & & \hi{क} & \ipa{/k/} & 44 \\
        \hi{ी} & \ipa{/iː/} & \del & & 76 & \hi{ब} & \ipa{/b/} & \hi{भ} & \ipa{/bʱ/} & 42 \\
        \hi{य} & \ipa{/j/} & \del & & 76 & \hi{र} & \ipa{/r/} & \del & & 42 \\
        \bottomrule
    \end{tabular}
    \caption{Most frequent transcription errors for Garhwali ASR. Key: \spc{} = space; $\epsilon$ = empty string; Deva = Devanagari script.}
    \label{tab:misclassifications}
\end{table}

\paragraph{Character-Level Error Patterns}
We further analyze character-level errors produced by \texttt{w2vBERT} fine-tuned on Garhwali. Table~\ref{tab:misclassifications} summarizes the most frequent substitutions, insertions, and deletions, which can be grouped into three broad categories: (1) word-boundary errors (e.g., space versus deletion), (2) vowel and consonant length confusions (e.g., \hi{ु} vs.\ \hi{ू}), and (3) aspiration-related errors (e.g., \hi{ब} vs.\ \hi{भ}).

Word-boundary errors are the most frequent, with 438 instances of omitted spaces and 289 instances of spurious space insertions, contributing to the observed WER despite relatively lower CER. The next most common errors involve vowel length and halant usage. In Garhwali, consonant-final words and consonant clipping, often marked using the halant \hi{्} , are common, whereas Hindi typically enforces an inherent vowel at word endings. Models pre-trained on Hindi orthographic conventions therefore tend to regularize Garhwali forms toward Hindi norms, resulting in systematic deletion or insertion of the halant. These error patterns are consistent with phonological and orthographic differences between Hindi and Garhwali.

\section{Conclusion}
In this work, we present a comprehensive study of dialectal speech recognition across a diverse set of Devanagari-script Indic language varieties, with a particular focus on understanding how dialectal variation interacts with cross-lingual transfer in low-resource ASR. Through a large-scale empirical evaluation, we find that while ASR performance is generally associated with phylogenetic distance across languages, this factor alone does not explain performance in the dialectal setting. In particular, when evaluating on dialects in the zero-shot setting, we observe lower word error rates when the fine-tuning language is a dialect or nonstandard variety. In many cases, fine-tuning on small amounts of dialectal speech yields performance comparable to or better than fine-tuning on larger amounts of phylogenetically closer, high-resource standardized languages.

Across multiple phylogenetic subgroups, our results consistently demonstrate that including higher-resource mainstream languages during fine-tuning does not reliably improve zero-shot ASR performance on dialectal evaluation sets. Instead, whether the fine-tuning data itself reflects dialectal speech emerges as a more informative predictor of performance than phylogenetic proximity alone. These findings highlight the importance of treating dialects as distinct acoustic and linguistic entities rather than as minor variants of standardized languages when designing ASR systems.

We further present the first detailed ASR analysis for Garhwali, a nonstandard Pahari language variety, and show that a \texttt{w2vBERT}-based model fine-tuned on Garhwali achieves the best performance among the evaluated architectures. Although the resulting word error rate of 49.3\% remains insufficient for fully automated transcription, this case study illustrates both the challenges of dialectal ASR and the benefits of dialect-specific modeling. Our quantitative error analysis further reveals substantial bias toward Hindi in both multilingual and Hindi-fine-tuned models, manifesting in systematic normalization of dialectal and code-mixed forms, and underscoring the need for dialect-aware data selection and modeling strategies in future ASR systems.

Overall, our findings suggest that effective ASR for low-resource dialects requires moving beyond default assumptions of phylogenetic similarity and toward evaluation and modeling practices that explicitly account for dialectal variation.

\section*{Limitations}
Our study is based on the VAANI dataset, which contains varying amounts of data across language varieties. Although this diversity allows us to evaluate ASR performance across a wide range of realistic dialectal settings, differences in dataset size may influence performance comparisons across languages. We mitigate this effect where possible by controlling the amount of fine-tuning data used across languages, though some variability remains inherent to the dataset.

In addition, VAANI consists of spontaneous and naturally-occurring speech collected across diverse regions. While this enables evaluation under realistic acoustic and conversational conditions, the presence of background noise, disfluencies, and region-specific recording environments may introduce additional variability in model performance.

Our analysis is limited to Indic dialects and language varieties written in the Devanagari script. Although this choice allows for controlled comparisons within a shared orthographic system, it excludes Indic languages written in other scripts, and our findings may not directly generalize beyond the Devanagari-script subset.

Finally, our experiments focus on a specific set of self-supervised ASR architectures and fine-tuning strategies. While these models are representative of widely used contemporary approaches, different architectures or training objectives may exhibit different transfer behaviors. 

\bibliography{references}

\appendix
\onecolumn
\section{Appendix}
\subsection{Baseline Evaluation}
We used the \texttt{IndicWav2Vec-Hindi} model to establish a baseline for the performance of off-the-shelf models on the languages in the VAANI dataset.  We further divided the results into two tables based on whether the Evaluation Language was seen by the \texttt{IndicWav2Vec-Hindi} model during pretraining or not.

\begin{table}[!htp]
\centering
\small
\setlength{\tabcolsep}{6pt}
\begin{tabular}{lcc}
\toprule
Language & CER & WER \\
\midrule
Bhojpuri      & 0.786 & 0.732 \\
Chhattisgarhi & 0.783 & 0.757 \\
Hindi         & 0.752 & 0.504 \\
Konkani       & 0.851 & 0.916 \\
Maithili      & 0.763 & 0.696 \\
Marathi       & 0.821 & 0.870 \\
Nepali        & 0.785 & 0.957 \\
Rajasthani    & 0.768 & 0.762 \\
\bottomrule
\end{tabular}
\caption{IndicWav2Vec-Hindi performance on languages seen during pre-training.}
\label{tab:seen_languages_split}
\end{table}

\begin{table}[!htp]
\centering
\small
\setlength{\tabcolsep}{4pt}
\begin{tabular}{lcc @{\hspace{6pt}} lcc}
\toprule
Language & CER & WER & Language & CER & WER \\
\midrule
Angika    & 0.791 & 0.725 & Khortha   & 0.829 & 0.743 \\
Awadhi    & 0.774 & 0.771 & Kumaoni   & 0.759 & 0.596 \\
Bajjika   & 0.784 & 0.784 & Kurukh    & 0.825 & 0.875 \\
Bhili     & 0.860 & 0.944 & Magadhi   & 0.782 & 0.700 \\
Bundeli   & 0.774 & 0.646 & Malvani   & 0.808 & 0.881 \\
Garhwali  & 0.760 & 0.820 & Marwari   & 0.752 & 0.740 \\
Gondi     & 0.718 & 0.945 & Nagpuri   & 0.797 & 0.834 \\
Halbi     & 0.788 & 0.873 & Sadri     & 0.742 & 0.738 \\
Haryanvi  & 0.811 & 0.766 & Surgujia  & 0.762 & 0.747 \\
Jaipuri   & 0.696 & 0.532 & Surjapuri & 0.829 & 0.880 \\
Khariboli & 0.784 & 0.605 & Thethi    & 0.796 & 0.707 \\
\bottomrule
\end{tabular}
\caption{IndicWav2Vec-Hindi performance on related but unseen languages.}
\label{tab:unseen_languages_split}
\end{table}

\subsection{Orthographic Analysis}
We computed the total number of words (tokens), total number of unique words (types) and the total number of words that only appear once (Hapax) per all the languages in the VAANI dataset. We also compute the total percentage of words that occur only once (Hapax\%) and the ratio of unique words to total number of words (TTR) in the table \ref{tab:hapax_analysis}. 

\begin{table}[!htp]
\centering
\small
\begin{tabular}{lrrrrr}
\toprule
     Language &  Tokens &  Types & Hapax & Hapax \% &    TTR \\
\midrule
        Gondi &   3,136 &  1,562 & 1,168 &    74.78 & 0.4981 \\
       Thethi &   1,556 &    654 &   469 &    71.71 & 0.4203 \\
        Bhili &   1,420 &    632 &   448 &    70.89 & 0.4451 \\
       Kurukh &   2,022 &    798 &   559 &    70.05 & 0.3947 \\
       Nepali &   2,169 &    776 &   536 &    69.07 & 0.3578 \\
    Surjapuri &   2,450 &    860 &   593 &    68.95 & 0.3510 \\
        Sadri &   6,599 &  2,096 & 1,417 &    67.60 & 0.3176 \\
      Nagpuri &   1,571 &    612 &   404 &    66.01 & 0.3896 \\
      Malvani &   9,123 &  2,501 & 1,636 &    65.41 & 0.2741 \\
      Konkani &  31,668 &  6,327 & 4,055 &    64.09 & 0.1998 \\
       Awadhi &   2,346 &    887 &   561 &    63.25 & 0.3781 \\
     Surgujia &   5,611 &  1,524 &   950 &    62.34 & 0.2716 \\
      Bundeli &   7,734 &  1,613 &   969 &    60.07 & 0.2086 \\
    Khariboli &  15,345 &  2,465 & 1,476 &    59.88 & 0.1606 \\
       Angika &  34,348 &  5,420 & 3,233 &    59.65 & 0.1578 \\
     Garhwali &  77,030 & 10,431 & 6,216 &    59.59 & 0.1354 \\
     Haryanvi &   3,747 &    983 &   584 &    59.41 & 0.2623 \\
        Halbi &  16,716 &  3,351 & 1,979 &    59.06 & 0.2005 \\
      Marathi & 171,424 & 16,876 & 9,919 &    58.78 & 0.0984 \\
      Bajjika &  32,049 &  4,681 & 2,717 &    58.04 & 0.1461 \\
      Magadhi &  55,891 &  7,101 & 4,093 &    57.64 & 0.1271 \\
      Khortha &  38,324 &  4,499 & 2,584 &    57.43 & 0.1174 \\
      Jaipuri &   2,074 &    599 &   339 &    56.59 & 0.2888 \\
Chhattisgarhi & 169,861 & 13,024 & 7,287 &    55.95 & 0.0767 \\
     Bhojpuri & 209,523 & 15,243 & 8,508 &    55.82 & 0.0728 \\
      Marwari &  85,883 &  8,279 & 4,582 &    55.34 & 0.0964 \\
      Kumaoni &  22,911 &  3,190 & 1,737 &    54.45 & 0.1392 \\
     Maithili & 194,541 & 14,317 & 7,779 &    54.33 & 0.0736 \\
        Hindi & 156,389 &  8,156 & 4,381 &    53.72 & 0.0522 \\
   Rajasthani & 116,180 &  7,988 & 4,273 &    53.49 & 0.0688 \\
\bottomrule
\end{tabular}
\caption{Hapax-Legomena (unique words) per training set.}
\label{tab:hapax_analysis}
\end{table}

\subsection{Garhwali ASR}

We evaluated multiple model architecture varieties on the Garhwali language test split from the VAANI dataset. For each model architecture, we first evaluated the performance without finetuning first and then after finetuning on the Training split of the Garhwali dataset. For each finetuning experiment except for the \texttt{Whisper} model, we generated the vocab of characters present in the training and validation split. 

\begin{table}[!htp]
    \centering
    
    \resizebox{\textwidth}{!}{%
        \begin{tabular}{lcccccccccc}
            \toprule
            Model Type & Tr. Lang & FT Lang & Gen Vocab? & Metric & Tr. Error & Val Loss & Tr. Loss & Steps & Test WER & Test CER \\
            \midrule
            wav2vec2CTC & - & Garhwali & Yes & WER & 0.769 & 1.576 & 0.403 & 6500 & 0.769 & - \\
            wav2vec2CTC & - & - & Yes & - & - & - & - & - & 1.015 & 2.436 \\
            \midrule
            XLS-R & - & Garhwali & Yes & WER & 0.735 & 1.554 & 0.171 & 1600 & 0.650 & 0.270 \\
            XLS-R & - & - & Yes & - & - & - & - & - & 1.000 & 1.292 \\
            \midrule
            wav2vec2 BERT & - & Garhwali & Yes & CER & 0.197 & 0.940 & 0.397 & 1200 & \textbf{0.493} & \textbf{0.193} \\
            wav2vec2 BERT & - & - & Yes & - & - & - & - & - & 1.000 & 1.600 \\
            \midrule
            Whisper-small & Hindi & Garhwali & No & CER & 0.242 & 0.974 & 0.001 & 4000 & 0.629 & 0.650 \\
            Whisper-small & Hindi & - & No & - & - & - & - & - & 2.518 & 1.262 \\
            \midrule
            HuBERT & - & Garhwali & Yes & CER & - & - & - & - & 0.515 & 0.199 \\
            \bottomrule
        \end{tabular}%
    }
    \caption{Comparison of Speech Models (Garhwali Fine-tuning vs. Baselines)}
    \label{tab:app_garhwali_models_performance}
\end{table}

\end{document}